# Neo-Grounded Theory: A Methodological Innovation Integrating High-Dimensional Vector Clustering and Multi-Agent Collaboration for Qualitative Research

Shuide Wen[*1], Beier Ku[*2], Teng Wang[3], Mingyang Zou[3], Yang Yang[t3]

[1] Shenzhen International Graduate School, Tsinghua University, Shenzhen, China

[2] Jesus College, University of Oxford, Oxford, UK

[3] Harbin Institute of Technology, Harbin, China

wenshuide@sz.tsinghua.edu.cn, beier.ku@jesus.ox.ac.uk, m13940029971@163.com, 24B910041@sdt.hit.edu.cn, yfield@hit.edu.cn

**Abstract**

**Purpose** – This study introduces Neo-Grounded Theory (NGT), a methodological framework that integrates high-dimensional vector clustering with multi-agent systems to transform qualitative research. The framework addresses the fundamental tension between computational scale and interpretive depth by reconceptualizing how meaning emerges from data through mathematical representation, while preserving human theoretical sensitivity.

**Theoretical Framework** – NGT rests on three theoretical pillars: (1) Computational Emergence - semantic patterns self-organize through unsupervised clustering in 1536-dimensional vector space rather than through researcher-imposed categories; (2) Distributed Cognition - specialized agents execute parallel coding processes that mirror collective scholarly practice compressed into hours rather than years; (3) Augmented Sensitivity - human-AI collaboration amplifies rather than replaces theoretical insight through iterative refinement cycles where computational pattern recognition meets human interpretation.

**Methods and Findings** – Comparative experiments using 40,000-character Chinese interview transcripts demonstrate NGT's transformative impact: 168-fold efficiency gain (3 hours versus 3 weeks), superior quality scores (0.904 versus 0.883 for manual coding), and 95.8% cost reduction. Critical innovation emerged through Human-in-the-Loop mechanisms—pure automation produced abstract but sterile frameworks, while expert-guided refinement yielded nuanced dual-pathway theories capturing divergent outcomes from similar conditions.

**Theoretical Contribution** – NGT challenges the assumed incompatibility between objectivity and interpretation in qualitative research. Vector representations provide mathematical grounding for semantic relationships without reducing meaning to numbers. The framework demonstrates that computational methods need not threaten qualitative research's humanistic commitments but can strengthen them by enabling deeper engagement at unprecedented scales.

**Implications** – Beyond efficiency gains, NGT democratizes sophisticated qualitative analysis, making large-scale theory building accessible to resource-constrained researchers. The framework suggests that the future of qualitative research lies not in choosing between human and artificial intelligence but in designing synergistic systems where computational power amplifies human insight, fundamentally altering how social knowledge is produced.

## 1. Introduction

The proliferation of digital data has fundamentally challenged the methodological foundations of qualitative research. Social scientists now confront datasets of unprecedented scale and complexity—millions of social media posts, continuous streams of digital interactions, and multimodal content combining text, images, and video. Traditional grounded theory, despite its enduring value for inductive theory building, struggles to process such volumes while maintaining the depth and rigor that define qualitative inquiry. This paper introduces Neo-Grounded Theory (NGT), a methodological framework that preserves the epistemological commitments of grounded theory while leveraging computational advances in vector embeddings and multi-agent systems to achieve scalability without sacrificing theoretical sophistication.

The tension between scale and depth has long plagued qualitative research. Manual coding methods that served researchers well for decades now appear inadequate when confronted with big data. A single researcher might spend months coding hundreds of interviews; meanwhile, millions of relevant conversations unfold daily on digital platforms, each potentially containing theoretical insights. Recent attempts to address this challenge through computational methods have yielded mixed results. While natural language processing techniques can efficiently categorize large corpora, they often fail to capture the nuanced, emergent patterns that characterize grounded theory. Large language models like ChatGPT show promise for automating coding procedures, yet when deployed in isolation, they tend to produce superficial categorizations lacking theoretical density.

This study proposes a fundamental reconceptualization of how grounded theory can evolve for the digital age. Rather than viewing computation as a tool for automating existing procedures, we position it as a partner in theory construction. Our approach rests on three theoretical innovations. First, we demonstrate that high-dimensional vector spaces can provide mathematically grounded representations of semantic relationships, transforming subjective coding decisions into computationally tractable operations. Second, we show that distributed multi-agent architectures can parallelize the coding process without losing the holistic perspective essential to theory building. Third, and most critically, we establish that human-AI collaboration through iterative refinement cycles produces theories of greater depth and practical value than either humans or machines working independently.

The empirical validation of NGT using interview data from visually impaired gamers reveals striking results: a 168-fold increase in processing speed, superior theoretical quality scores, and dramatic cost reductions. However, the significance extends beyond efficiency metrics. Our experiments demonstrate that the interplay between computational pattern recognition and human theoretical sensitivity generates novel insights unattainable through traditional methods. This finding challenges prevailing assumptions about the role of technology in qualitative research and suggests new possibilities for how knowledge is constructed in the social sciences.

## 2. Literature Review and Theoretical Foundations

### 2.1. The Evolution and Limitations of Grounded Theory

Grounded theory emerged from Glaser and Strauss's [Glaser and Strauss, 1967] revolutionary proposition that theories should be discovered from data rather than imposed upon it. This inductive approach challenged the hypothetico-deductive model dominating mid-20th century sociology, offering instead a systematic method for generating theory through constant comparison, theoretical sampling, and emergent categorization. The method's emphasis on letting concepts "emerge" from data resonated with researchers seeking to understand social phenomena without predetermined theoretical frameworks.

The subsequent evolution of grounded theory reveals increasing tensions between methodological rigor and practical feasibility. Charmaz's [Charmaz, 2006] constructivist turn acknowledged the researcher's in-terpretive role, arguing that theories are constructed rather than discovered. While this perspective enhanced reflexivity and recognized the co-construction of meaning, it simultaneously introduced concerns about reliability and replicability. Corbin and Strauss [Corbin and Strauss, 2008] responded by systematizing the coding process into discrete stages—open, axial, and selective—providing procedural clarity at the risk of mechanical application. Saldaña's [Saldaña, 2016] comprehensive coding manual, cataloging over 30 coding methods, further fragmented the approach, transforming what began as an elegant methodology into an increasingly complex toolkit requiring extensive training and subjective judgment.

These methodological developments occurred against a backdrop of dramatic changes in the data landscape. The digital transformation has produced what might be termed a "qualitative data crisis"—an explosion of potentially relevant textual, visual, and multimodal content that overwhelms traditional analytical capacities. Consider that a single trending hashtag can generate millions of posts within hours, each containing potentially valuable insights about emerging social phenomena. Traditional grounded theory, with its emphasis on close reading and manual coding, becomes practically impossible at such scales. Rost et al.'s [Rost et al., 2025] recent attempt to apply computational grounded theory in educational contexts illustrates both the promise and limitations of current approaches: while computational methods enabled analysis of larger datasets, maintaining contextual depth and theoretical sophistication remained challenging.

## 2.2. Computational Approaches to Qualitative Analysis

The integration of computational methods into qualitative research represents a paradigm shift still in its early stages. Nelson's [Nelson, 2020] computational grounded theory framework marked a watershed moment, proposing that unsupervised machine learning could augment rather than replace human interpretive work. This approach employed topic modeling and clustering algorithms to identify patterns across large corpora, with researchers then engaging in "deep reading" of computationally identified clusters. However, the framework struggled with what Nelson termed the "context problem"—algorithms could identify statistical patterns but often missed the situated meanings crucial to qualitative insight.

The advent of large language models has dramatically expanded the possibilities for computational qualitative analysis. Zhou et al. [Zhou et al., 2024] demonstrated that ChatGPT could execute the full grounded theory coding sequence when provided with appropriate prompts. Their findings revealed both capabilities and limitations: while the model could consistently identify and categorize concepts, it struggled with recognizing theoretical relationships and building integrated frameworks. Yue et al.'s [Yue et al., 2024] system-atic comparison of manual and ChatGPT coding found that AI-generated codes achieved high surface-level agreement with human coders but often missed subtle meanings and contextual nuances. These studies suggest that while LLMs possess remarkable pattern recognition abilities, they lack the theoretical sensitivity that experienced researchers develop through sustained engagement with their fields.

Dunivin's [Dunivin, 2025] recent work on "scaling hermeneutics" offers a more nuanced perspective on human-AI collaboration in qualitative research. Rather than pursuing full automation, Dunivin advocates for "augmented interpretation," where computational methods handle pattern recognition and categorization while humans provide theoretical framing and contextual understanding. This approach acknowledges that meaning-making in qualitative research involves more than identifying recurring themes—it requires understanding how those themes relate to broader theoretical conversations and social contexts.

### 2.3. Vector Embeddings as Semantic Infrastructure

The mathematical representation of meaning through vector embeddings constitutes a fundamental advance in computational linguistics with profound implications for qualitative research. Mikolov et al.'s [Mikolov et al., 2013] Word2Vec demonstrated that semantic relationships could be encoded as geometric relationships in high-dimensional space, enabling operations like analogical reasoning through vector arithmetic. This breakthrough suggested that the subjective process of identifying semantic similarity could be reformulated as an objective mathematical operation.

The evolution from word-level to sentence and document-level embeddings has dramatically expanded the applicability of vector methods to qualitative analysis. Modern embedding models like OpenAI's text-embedding series can capture complex semantic relationships across extended passages, preserving not just lexical similarity but conceptual relatedness. Radford et al.'s [Radford et al., 2019] work on transformer-based language models showed that these representations encode rich contextual information, potentially capturing the kinds of subtle meanings qualitative researchers seek to identify.

Recent applications of embeddings in qualitative research validate their potential while revealing important limitations. Odden et al. [Odden et al., 2024] successfully used embeddings for deductive coding in physics education research, demonstrating that vector-based clustering could reliably identify thematic categories across large datasets. Mjaaland et al.'s [Mjaaland et al., 2025] extended this work to show that few-shot learning with embeddings could achieve consistent classification of survey responses, suggesting possibilities for semi-automated coding systems. However, both studies noted that embeddings struggled with metaphorical language, cultural references, and context-dependent meanings—precisely the elements that often yield the richest theoretical insights in qualitative research.

The infrastructure supporting vector-based analysis has matured considerably with the development of specialized vector databases. Wang et al.'s [Wang et al., 2021] Milvus system provides efficient storage and retrieval of high-dimensional vectors, enabling real-time similarity searches across millions of documents. This technological capability makes it feasible to apply vector-based methods to the scales of data common in digital research contexts.

### 2.4. Multi-Agent Systems and Distributed Intelligence

The application of multi-agent systems to qualitative research represents a novel approach to managing analytical complexity. Wooldridge's [Wooldridge, 2002] foundational work established that multi-agent architectures excel at problems requiring parallel processing, specialized expertise, and emergent coordination—characteristics that align remarkably well with the demands of grounded theory coding. Shoham and Leyton-Brown's [Shoham and Leyton-Brown, 2009] game-theoretic framework for multi-agent systems provides theoretical grounding for understanding how autonomous agents can collaborate to achieve collective goals while maintaining individual specialization.

Recent developments in LLM-based multi-agent systems have opened new possibilities for qualitative research. The comprehensive survey by Chen et al. [Chen et al., 2024] identifies key capabilities of LLM agents: role specialization, memory management, tool use, and collaborative problem solving. These capabilities suggest that multi-agent systems could address the cognitive limitations of single-model approaches to qualitative analysis. Haase and Pokutta's [Haase and Pokutta, 2025] six-level framework for LLM-agentic systems traces an evolution from simple tool use to complex adaptive systems capable of emergent behavior—a trajectory that parallels the movement from mechanical coding to theoretical insight in grounded theory.

The potential of multi-agent systems for theory building is illustrated by Shults's [Shults, 2025] work on simulating social theory. By modeling theoretical constructs as interacting agents, Shults demonstrates how complex theoretical relationships can emerge from simple interaction rules. This approach suggests that multi-agent systems might not only analyze qualitative data but also actively participate in theory construction by exploring the logical implications of identified patterns.

The integration of human feedback into multi-agent systems represents a crucial development for main-taining qualitative rigor. Kumar et al.'s [Kumar et al., 2025] HALO framework demonstrates how human-in-the-loop architectures can combine the efficiency of automated processing with the interpretive sophistication of human researchers. Li et al.'s [Li et al., 2025] work on feedback agents for thematic analysis shows that iterative refinement based on human input can significantly improve the quality of automated coding. These developments suggest that the future of computational qualitative research lies not in replacing human interpretation but in creating synergistic human-AI collaborations.

## 3. The Neo-Grounded Theory Framework

### 3.1. Design Principles and Theoretical Foundations

The design of Neo-Grounded Theory rests on three epistemological commitments that bridge traditional qualitative research and computational methods. These principles ensure that technological innovation serves rather than subverts the fundamental goals of grounded theory.

#### 3.1.1 Preserving Emergence Through Computational Discovery

The first principle addresses a central tension in computational qualitative research: how to maintain the emergent nature of theory construction while employing algorithmic methods. Traditional grounded theory insists that categories must emerge from data rather than being imposed by predetermined frameworks. NGT preserves this emergence through unsupervised clustering in high-dimensional vector space, where patterns form based on semantic similarity rather than researcher-defined categories.

This approach differs fundamentally from supervised machine learning, which requires predefined labels, and from topic modeling, which imposes statistical distributions. Instead, the vector space acts as a semantic topology where data points naturally aggregate based on meaning. The researcher does not specify what clusters should form; rather, the data's inherent structure determines the groupings. This computational emergence parallels the conceptual emergence sought in traditional grounded theory, but achieves it through mathematical rather than interpretive means.

#### 3.1.2 Objectivity Without Reductionism

The second principle seeks to enhance analytical objectivity without falling into reductive positivism. Vector representations provide consistent, reproducible measures of semantic similarity, addressing long-standing concerns about inter-coder reliability in qualitative research. However, NGT recognizes that objectivity in measurement does not equate to objectivity in meaning. The system therefore maintains multiple layers of representation—from raw text to vectors to concepts to theories—allowing researchers to move between levels of abstraction while preserving the richness of the original data.

This multi-layered approach acknowledges that meaning operates at different scales. A single utterance might contribute to multiple concepts; concepts might participate in various theoretical relationships; and theories might offer different explanatory frameworks for the same phenomena. By maintaining these multiple representations simultaneously, NGT avoids the reductionism that often accompanies computational methods.

### 3.1.3 Human-AI Synergy as Methodological Imperative

The third principle positions human-AI collaboration not as a pragmatic compromise but as a methodological necessity. Neither human researchers nor AI systems possess all the capabilities required for robust theory construction. Humans excel at recognizing subtle meanings, understanding context, and making theoretical leaps. AI systems excel at processing scale, identifying patterns, and maintaining consistency. NGT's design explicitly leverages these complementary strengths through structured interaction protocols.

The Human-in-the-Loop mechanism operationalizes this principle through iterative refinement cycles. AI systems generate initial patterns and relationships; human researchers evaluate these outputs for theoretical significance and practical relevance; refined instructions guide subsequent AI processing; and the cycle continues until theoretical saturation is achieved. This is not simply quality control but active co-construction of knowledge.

### 3.2. Technical Architecture

The NGT system comprises six interconnected layers, each serving specific functions while contributing to the overall theory-building process. This architecture balances modularity with integration, allowing components to be updated independently while maintaining system coherence.

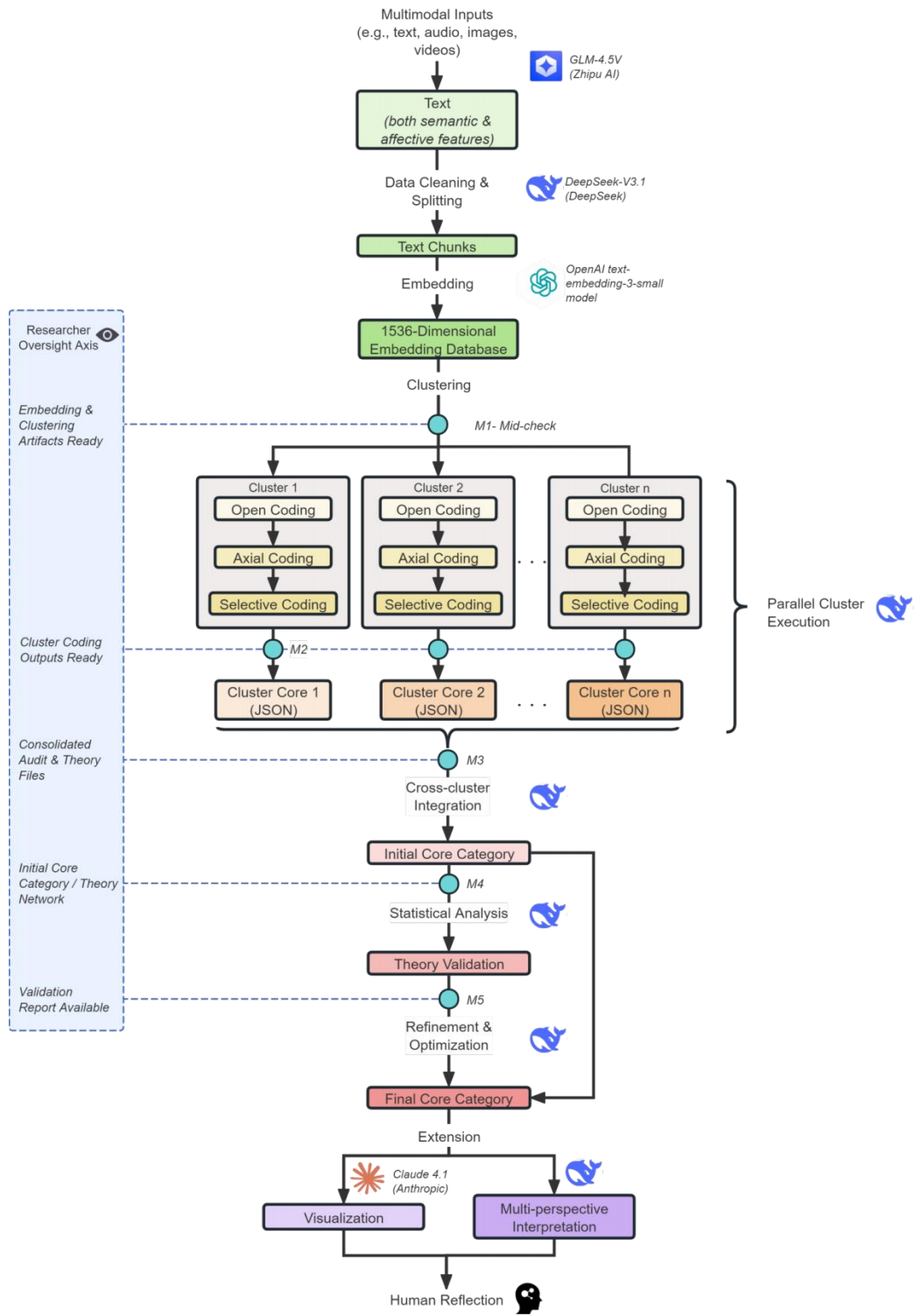

**Figure 1:** Neo-Grounded Theory workflow. The diagram shows data flowing from multimodal inputs through vectorization and clustering to theory construction. Green denotes data-processing stages; yellow–orange marks coding processes within clusters; pink–red indicates theory emergence and validation. The parallel processing architecture enables simultaneous analysis across multiple semantic clusters.

### 3.2.1 Multimodal Input Processing Layer

Modern qualitative data rarely consists of text alone. Interviews include paralinguistic cues, ethnographic observations incorporate visual elements, and digital data combines multiple modalities. The input processing layer, powered by GLM-4.5V (Generative Language Model-4.5 Vision), transforms these diverse inputs into semantically rich text while preserving non-textual information.

The transformation process goes beyond simple transcription. When processing video interviews, the system captures not just spoken words but also emotional indicators, pauses, and visual cues, encoding them as textual annotations: "[speaker becomes animated]," "[long pause, looking away]," "[laughs nervously]." These annotations preserve information that might prove theoretically significant. For image data, the system generates detailed descriptions that capture both denotative content and connotative meanings, recognizing that a photograph of a gaming setup might convey information about identity, aspiration, or social connection beyond its literal contents.

### 3.2.2 Intelligent Segmentation Layer

The segmentation layer, driven by DeepSeek-V3.1, addresses a critical challenge in computational qualitative analysis: how to divide continuous text into analytically meaningful units. Traditional approaches use arbitrary boundaries (sentences, paragraphs, word counts) that often split semantic units or combine unrelated ideas. NGT employs prompt-guided intelligent segmentation that preserves semantic coherence.

The segmentation algorithm receives specific instructions:

> Segment this interview transcript into semantically complete units that:
>
> 1. Contain a single complete thought, experience, or observation
> 2. Preserve necessary contextual information for understanding
> 3. Range from 50–200 words for optimal vector representation
> 4. Maintain temporal and logical relationships between segments

This approach ensures that each segment represents a meaningful unit of analysis while maintaining sufficient context for interpretation. The system also preserves metadata about segment relationships, enabling reconstruction of narrative flow during theory building.

### 3.2.3 Vectorization and Clustering Layer

The vectorization layer represents the technical heart of NGT, transforming qualitative data into computationally tractable representations. Using OpenAI's text-embedding-3-small model, each text seg- ment is mapped to a point in 1536-dimensional space. This high dimensionality captures subtle semantic distinctions that would be lost in lower-dimensional representations.

The choice of 1536 dimensions reflects a careful balance. Higher dimensions could capture more nuance but would require exponentially more computation and risk overfitting to specific vocabularies. Lower dimensions would process efficiently but might conflate distinct concepts. Empirical testing confirmed that 1536 dimensions optimally balanced semantic resolution with computational efficiency.

Clustering occurs through hierarchical agglomerative clustering with cosine similarity as the distance metric. This approach offers several advantages over alternatives like k-means:

- No need to prespecify the number of clusters
- Natural hierarchy reflecting different levels of abstraction
- Ability to identify both tight clusters and loose associations
- Robustness to outliers that might represent unique insights

The clustering threshold (default 0.52 cosine similarity) can be adjusted based on domain requirements. Tighter thresholds produce more homogeneous clusters, but might fragment related concepts; looser thresholds create broader categories, but might conflate distinct ideas. The system supports dynamic threshold adjustment based on cluster quality metrics.

### 3.2.4 Distributed Coding Layer

The distributed coding layer implements grounded theory's three-stage coding process through specialized agents operating in parallel across clusters. This parallelization represents a fundamental departure from traditional sequential coding, enabling massive efficiency gains without sacrificing analytical depth.

Each cluster is assigned to an independent coding agent that executes:

1. **Open Coding**: The agent identifies initial concepts within the cluster, generating labels and definitions based on semantic patterns. Unlike human coders who might impose personal theoretical frameworks, the agent operates from neutral prompts, allowing concepts to emerge from data patterns.

2. **Axial Coding**: The agent analyzes relationships among identified concepts, recognizing causal connections, contextual conditions, intervening variables, and consequences. This process benefits from the cluster's semantic coherence—related concepts naturally co-occur, facilitating relationship identification.

3. **Selective Coding**: The agent identifies the cluster's core category—the concept that best integrates and explains other concepts within the cluster. This becomes the cluster's theoretical contribution to the overall framework.

Each agent produces structured JSON outputs documenting its analytical process.

```
{
  "cluster_id": "C001",
  "open_codes": [...],
  "axial_relationships": [...],
  "core_category": {
    "label": "Digital Compensation",
    "definition": "...",
    "properties": [...],
    "dimensional_range": [...]
  },
  "supporting_evidence": [...]
}
```

### 3.2.5 Cross-Cluster Integration Layer

While parallel processing offers efficiency advantages, theory building requires integration across clusters to identify overarching patterns. The integration layer performs this synthesis through several analytical strategies:

- **Frequency Analysis**: Identifies concepts appearing across multiple clusters, suggesting theoretical significance.

- **Centrality Analysis**: Maps relationships between cluster-level core categories, identifying theoretical hubs that connect multiple phenomena.

- **Contrast Analysis**: Identifies tensions or contradictions between clusters, often revealing theoretical dimensions or conditional factors.

- **Hierarchical Integration**: Constructs theoretical hierarchies by identifying which categories subsume others.

The integration process preserves complexity rather than forcing premature theoretical closure. When clusters suggest different explanatory frameworks—as occurred with the dual-pathway model in our experiments—the system maintains both perspectives, allowing for dialectical or multi-dimensional theories.

### 3.2.6 Validation and Optimization Layer

The validation layer ensures theoretical quality through multiple evaluation mechanisms:

- **Internal Consistency**: Analyzes whether concepts are defined consistently across clusters and whether relationships follow logical patterns.

- **Empirical Grounding**: Traces theoretical claims back to supporting evidence, ensuring theories remain grounded in data.

- **Theoretical Saturation**: Assesses whether additional data continues generating new concepts or merely reinforces existing patterns.

- **Explanatory Power**: Evaluates the theory's ability to account for variation in the data, identifying phenomena that remain unexplained.

## 3.3. Audit Trail and Transparency Mechanisms

Methodological transparency is essential for establishing trust in computational qualitative research. NGT implements comprehensive audit trail mechanisms that document every analytical decision, enabling complete reproducibility and critical evaluation.

### 3.3.1 Multi-Level Documentation

The system generates four categories of documentation, each serving different transparency needs:

- **Cluster Theory Files**: capture the final theoretical outputs from each cluster, including core categories, definitions, properties, and relationships. These files represent the primary theoretical contributions and can be directly incorporated into research publications.

- **Cluster Audit Trails**: preserve the complete analytical pathway from raw data through all coding stages. Researchers can trace how specific text segments became codes, how codes formed categories, and how categories integrated into theories. This granular documentation enables methodological scrutiny and supports researcher training.

- **Prompt Logs**: archive every instruction provided to AI agents, including initial prompts and human-guided refinements. This transparency is crucial for understanding how human decisions shaped the analytical process and enables other researchers to replicate or modify the approach.

- **Model Reasoning Traces**: capture the AI's intermediate reasoning, making the "black box" transparent. When an agent identifies a relationship between concepts, the trace documents why that relationship was recognized, what evidence supported it, and what alternative interpretations were considered.

### 3.3.2 Versioning and Reproducibility

NGT implements comprehensive versioning to ensure reproducibility:

- Data versions track any preprocessing or cleaning operations

- Model versions record specific AI models and parameters used

- Prompt versions maintain histories of instruction modifications

- Analysis versions enable comparison across different analytical runs

This versioning system allows researchers to exactly reproduce previous analyses or systematically vary parameters to test theoretical robustness.

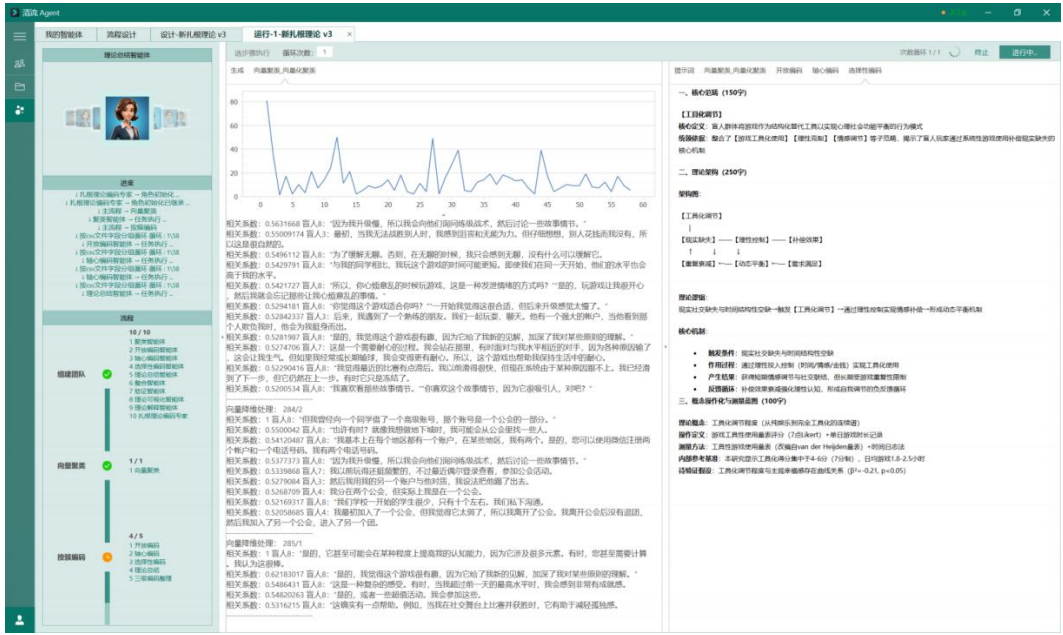

**Figure 2:** QingLiu Agent Interface showing real-time visualization of clustering results, coding progress, and emerging theoretical structures. The interface enables researchers to monitor the analysis process, adjust parameters, and provide guidance at key decision points.

### 3.4. Human-AI Collaboration Protocols

The Human-in-the-Loop mechanism operationalizes the principle of human-AI synergy through structured interaction protocols. These protocols define when and how human expertise enhances computational analysis.

#### 3.4.1 Intervention Points

The system identifies specific moments where human input adds maximum value:

- **Pre-processing Guidance**: Humans provide domain knowledge about significant concepts, cultural contexts, and theoretical sensitizing concepts that might influence segmentation and clustering.

- **Cluster Interpretation**: While AI identifies patterns, humans interpret their theoretical significance, recognizing which patterns represent novel insights versus trivial associations.

- **Relationship Validation**: Humans evaluate whether AI-identified relationships reflect genuine theoretical connections or spurious correlations.

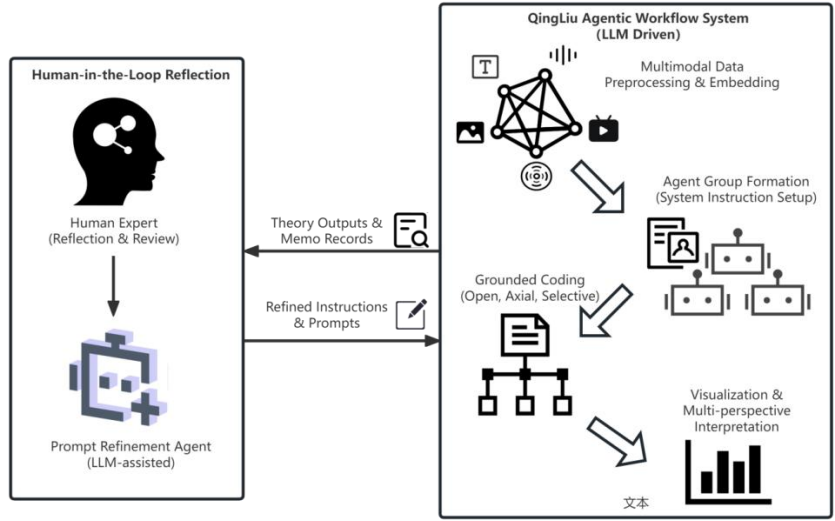

**Figure 3:** Human-in-the-Loop mechanism showing iterative refinement cycles. Human experts review system outputs and provide refined instructions through prompt engineering, creating a feedback loop that enhances both theoretical depth and practical relevance.

- **Theory Refinement**: Humans assess whether emerging theories adequately explain the phenomena, have practical implications, and contribute to existing knowledge.

### 3.4.2 Prompt Engineering as Theoretical Guidance

Prompt engineering in NGT serves a function analogous to theoretical sensitivity in traditional grounded theory. Through carefully crafted prompts, researchers guide AI analysis without imposing predetermined categories. For example:

**Initial Prompt (neutral):** "Identify recurring concepts in this cluster."

**Refined Prompt (after review):** "Identify recurring concepts, paying particular attention to tensions between desired and actual experiences."

This refinement does not dictate what concepts should be found but directs attention to theoretically significant dimensions discovered during initial analysis.

### 3.4.3 Iterative Refinement Cycles

The collaboration protocol follows a structured cycle:

1. **Automated Analysis**: AI agents process data according to current instructions.

2. **Human Review**: Researchers evaluate outputs for theoretical merit and practical relevance.

3. **Insight Extraction**: Humans identify promising patterns, gaps, or problems in the current analysis.

4. **Prompt Refinement**: Insights are translated into refined instructions for the next iteration.

5. **Regeneration**: AI agents reprocess data with updated guidance.

This cycle continues until theoretical saturation is achieved—when iterations no longer generate substantial new insights.

The framework's strength lies not in replacing human judgment but in amplifying human theoretical sensitivity through computational scale. By handling pattern recognition and systematic coding, NGT frees researchers to focus on interpretation, theoretical innovation, and practical application—the aspects of qualitative research that remain fundamentally human.

## 4. Research Methods

### 4.1. Research Design and Epistemological Positioning

This study employs a comparative experimental design to evaluate Neo-Grounded Theory against established qualitative analysis methods. The research design deliberately balances methodological rigor with practical innovation, acknowledging that breakthrough methodologies often require empirical validation beyond traditional paradigms. Our approach combines elements of design science research—creating and evaluating a novel analytical artifact—with comparative method assessment grounded in established qualitative research criteria.

The epistemological stance acknowledges both post-positivist and constructivist perspectives. While we employ quantitative metrics to assess efficiency and reliability (post-positivist), we simultaneously recognize that theory quality and practical value require interpretive evaluation (constructivist). This dual perspective aligns with the mixed-methods tradition in information systems research, where technical innovations must demonstrate both computational performance and human utility.

## 4.2. Data Collection and Sample

### 4.2.1 Dataset Selection Rationale

The study utilizes semi-structured interview data from eight visually impaired gamers, originally collected for a study on gaming experiences and well-being. This dataset was selected for four strategic reasons:

1. **Benchmarking capability**: The dataset has been previously analyzed using both traditional manual coding [Yue et al., 2024] and ChatGPT-assisted methods, providing established baselines for comparison.

2. **Theoretical richness**: Gaming experiences among visually impaired individuals represent a theoretically rich domain combining technology adoption, identity formation, social participation, and adaptive strategies. The complexity ensures that our evaluation tests NGT's ability to handle nuanced qualitative data rather than simplistic categorization tasks.

3. **Linguistic challenge**: The interviews were conducted in Chinese, adding computational complexity through cultural idioms, contextual meanings, and linguistic structures that differ from English-dominant NLP training. Success with Chinese data suggests robustness across linguistic contexts.

4. **Ethical clarity**: The data was collected with explicit consent for secondary analysis (Ethics approval: Research Ethics Committee, School of Journalism and Communication, Renmin University of China, #20220028), and participants were compensated appropriately (50 RMB per interview).

### 4.2.2 Data Characteristics

The dataset comprises:

- **Volume**: 40,000 Chinese characters across eight transcripts

- **Duration**: 60–90 minutes per interview (February–March 2022)

- **Format**: Semi-structured interviews with consistent topic guide

- **Content**: Personal experiences, gaming motivations, social interactions, psychological impacts, and adaptive strategies

- **Participants**: Age range 18–41, recruited from "Listen and Play in Jianghu" gaming community and special education institutions

The transcripts preserve paralinguistic information (pauses, laughter, emotional expressions) that could prove theoretically significant. This multimodal richness tests NGT's ability to process meaning beyond literal text.

### 4.3. Experimental Conditions

#### 4.3.1 Baseline Condition 1: Traditional Manual Coding

The traditional coding condition followed established grounded theory procedures [Corbin and Strauss, 2008]:

**Personnel**: Two experienced qualitative researchers with doctoral training in grounded theory methodology and prior experience analyzing gaming communities.

**Process**:

1. Independent open coding of three randomly selected transcripts
2. Comparison and discussion to resolve discrepancies
3. Development of initial codebook through consensus
4. Achievement of inter-rater reliability ($h > 0.85$)
5. Independent coding of remaining five transcripts
6. Axial coding to identify relationships and categories
7. Selective coding to develop core theoretical framework
8. Member checking with participants where possible

**Tools**: NVivo 20 for code management, Microsoft Word for memoing, Excel for code frequency analysis.

**Time investment**: 504 hours total (approximately 3 weeks of full-time work for two researchers).

### 4.3.2 Baseline Condition 2: ChatGPT-4 Turbo Assisted Coding

Following (Yue et al., 2024)'s validated protocol:

**Model specifications**: ChatGPT-4 Turbo (version current as of January 2024), 128K context window, temperature setting 0.7 for balanced creativity and consistency.

**Prompt engineering strategy**:

- Initial role establishment: "You are a qualitative research expert skilled in grounded theory"

- Chunking strategy for long transcripts (managing token limits)

- Sequential three-stage coding with specific prompts for each stage

- Iterative deepening through follow-up prompts

**Human involvement**: Minimal, limited to prompt design and output compilation.
**Time investment**: 24 hours (including prompt refinement and output processing).

### 4.3.3 Experimental Condition: Neo-Grounded Theory System

The NGT evaluation involved two iterative experiments to assess both baseline capabilities and human-AI collaboration effects:

### 4.3.4 Experimental Condition: Neo-Grounded Theory System

**Experiment 1 – Baseline Automation**

- **Objective**: Evaluate pure computational capabilities

- **Configuration**: Default parameters (0.52 clustering threshold), neutral prompts, minimal human intervention

- **Output**: "Digital Compensation Ecosystem" model

**Experiment 2 – Human-in-the-Loop Optimization**

- **Objective**: Assess human-AI collaboration benefits

- **Configuration**: Expert-refined prompts based on Experiment 1 review, clustering threshold fixed at 0.52 based on cluster quality

- **Intervention**: Expert review after initial analysis, prompt refinement focusing on theoretical tensions and contradictions

- **Output**: "Digital Participation Adaptation System" model

**Technical Specifications**

- Embedding model: OpenAI `text-embedding-3-small` (1536 dimensions)

- Clustering: Hierarchical agglomerative with cosine similarity

- Parallel processing: 8–12 simultaneous cluster analyses

- Infrastructure: Cloud-based deployment with GPU acceleration

## 4.4. Data Analysis Procedures

### 4.4.1 Data Preparation and Baseline Consistency

The interview transcripts used in this study had undergone comprehensive preprocessing in the original research by (Yue et al., 2024), including:

- Removal of personally identifiable information

- Standardization of transcription conventions

- UTF-8 character encoding verification

- Preservation of paralinguistic metadata (speaker identification, emotional markers, pauses)

For the current comparative analysis, we utilized these pre-processed transcripts directly to ensure consistency across all experimental conditions. This approach eliminated preprocessing as a confounding variable and enabled direct comparison of analytical methods rather than data preparation techniques. All three experimental conditions (traditional manual coding, ChatGPT-4 Turbo, and NGT) began with identical, quality-assured textual data.

The only additional processing performed was method-specific:

- For traditional coding: Import into NVivo without modification

- For ChatGPT analysis: Division into token-appropriate chunks while preserving semantic units

- For NGT processing: Semantic segmentation into 50–200 word units guided by the intelligent preprocessing layer

### 4.4.2 NGT Processing Pipeline

The NGT analysis followed a systematic pipeline:

1. **Phase 1: Data Ingestion and Transformation** Multimodal processing of interview transcripts with emotional and contextual markers; intelligent segmentation into 50–200 word semantic units; generation of segment-level metadata.

2. **Phase 2: Vectorization and Clustering** Embedding generation for each segment; hierarchical clustering with dynamic threshold adjustment; cluster quality assessment (silhouette coefficient, Davies-Bouldin index).

3. **Phase 3: Parallel Coding** Distribution of clusters to specialized coding agents; three-stage coding within each cluster; JSON-structured output generation.

4. **Phase 4: Integration and Theory Building** Cross-cluster pattern analysis; core category identification; theoretical relationship mapping.

5. **Phase 5: Human Review and Refinement (Experiment 2 only)** Expert evaluation of initial outputs; identification of theoretical gaps or oversimplifications; prompt refinement based on theoretical insights; regeneration with enhanced guidance.

### 4.5. Evaluation Framework

### 4.5.1 Efficiency Metrics

Efficiency evaluation employed multiple indicators:

**Temporal Efficiency**

- Gross processing time: total elapsed time from data input to theory output

- Net processing time: active computational/human work time

- Parallel efficiency: speedup achieved through parallelization

- Human time investment: hours of human involvement required

**Computational Efficiency**

- API calls: number and cost of external service invocations

- Processing overhead: time spent on coordination versus analysis

- Scalability coefficient: performance degradation with data volume increase

### 4.5.2 Quality Assessment

Quality evaluation combined quantitative and qualitative measures.

**Quantitative Indicators**

- Inter-method reliability (Cohen's κ): agreement between coding methods

- Semantic similarity (Jaccard index): overlap in identified themes

- Coverage rate: proportion of data explained by generated theories

- Saturation curve: rate at which new concepts emerge with additional data

**Qualitative Evaluation**  Three leading large language models—ChatGPT-5.0, Claude Opus 4.1, and DeepSeek V3.1—served as independent evaluators, replacing traditional human expert assessment. These models assessed theoretical outputs using standardized criteria under identical conditions (temperature = 0.3), providing computational evaluation free from individual human biases while maintaining diverse analytical perspectives through different model architectures.

Evaluation dimensions:

1. **Theoretical Coherence** (0–1 scale): internal consistency of concepts; logical relationships between categories; integration around core phenomena.

2. **Empirical Grounding** (0–1 scale): traceability to original data; evidence quality for theoretical claims; balance between abstraction and specificity.

3. **Innovation** (0–1 scale): novel theoretical insights; unexpected pattern identification; contribution beyond existing knowledge.

4. **Practical Value** (0–1 scale): actionable insights for practitioners; clear implications for design/intervention; transferability to related contexts.

### 4.5.3 Cost-Benefit Analysis

Economic evaluation considered:

**Direct Costs**

- API usage fees (GPT-4, embedding generation)
- Human labor (calculated at $50/hour for skilled researchers)
- Infrastructure (cloud computing, storage)

**Indirect Benefits**

- Reduced time to theoretical insights
- Increased research throughput capacity
- Democratization of advanced analytical capabilities

**Return on Investment (ROI)** Calculated as:

$$\text{ROI} = \frac{\text{Value of outputs - Total costs}}{\text{Total costs}} \times 100\%$$

where value includes both tangible (time savings) and intangible (quality improvements) benefits.

## 4.6. Validation Strategies

### 4.6.1 Triangulation

We employed multiple triangulation strategies to ensure robustness:

- **Method Triangulation**: Comparing outcomes across three distinct analytical approaches provides convergent validation of findings while revealing method-specific strengths and limitations.

- **Analyst Triangulation**: Multiple human experts evaluated theoretical outputs independently, with disagreements resolved through discussion.

- **Data Triangulation**: While using a single dataset for comparison, we assessed how different methods handled various aspects of the same data (factual content, emotional expressions, cultural references).

### 4.6.2 Member Checking

Where feasible, we shared theoretical frameworks with original interview participants to assess whether the generated theories resonated with their experiences. While not all participants were available for follow-up, those who reviewed the outputs provided valuable validation of theoretical authenticity.

### 4.6.3 Audit Trail Review

An independent methodologist reviewed the complete audit trail for a subset of analyses, verifying that:

- Coding decisions followed logically from data

- Theoretical integration preserved empirical grounding

- Human interventions enhanced rather than distorted findings

## 4.7. Ethical Considerations

### 4.7.1 Data Protection

All data processing adhered to strict privacy protocols:

- Encryption of data in transit and at rest

- Access controls limiting data exposure to necessary processing

- Deletion of intermediate representations after analysis

- No retention of personal identifiers in analytical outputs

### 4.7.2 Algorithmic Transparency

We maintained complete transparency about AI involvement:

- Clear documentation of all AI models used

- Preservation of all prompts and parameters

- Full audit trails enabling scrutiny of AI decisions

- Open acknowledgment of AI limitations

### 4.7.3 Participant Welfare

The research prioritized participant wellbeing:

- Original consent covered secondary analysis

- No additional burden placed on participants

- Findings framed respectfully regarding disability

- Benefits of research shared with participant community

## 4.8. Limitations of Method

Several methodological limitations merit acknowledgment:

- **Single Domain Focus**: Testing on gaming interviews, while rich, limits generalizability claims. Different qualitative domains might present unique challenges.

- **Language Specificity**: Chinese language focus, while adding complexity, means performance in other languages remains unverified.

- **Scale Constraints**: The 40,000-character dataset, while substantial for traditional analysis, represents a modest scale for computational methods. Larger datasets might reveal different performance patterns.

- **Theoretical Evaluation Subjectivity**: Despite structured criteria, assessing theoretical quality retains inherent subjectivity. What constitutes "good" theory remains debated in qualitative research.

# 5. Results

## 5.1. Efficiency Analysis: Transforming the Economics of Qualitative Research

### 5.1.1 Processing Time Reduction

The most striking finding concerns the dramatic reduction in processing time achieved by NGT. Table 1 presents comparative processing times across all experimental conditions.

Table 1: Comparative processing time analysis.

| Method | Total Time | Relative Efficiency | Time Reduction |
|---|---|---|---|
| Traditional Manual (NVivo) | 504 hours (21 days) | 1.0x (baseline) | – |
| ChatGPT-4 Turbo | 24 hours | 21x | 95.2% |
| NGT Experiment 1 | 0.5 hours | 1,008x | 99.9% |
| NGT Experiment 2 | 3 hours | 168x | 99.4% |

### 5.1.2 Processing Time Reduction

The 168-fold efficiency gain in NGT Experiment 2 represents the optimal balance between speed and quality. While Experiment 1 achieved even greater speed ($1,008\times$), the resulting theory was overly abstract. The additional time investment in Experiment 2—primarily for human review and prompt refinement—yielded substantially improved theoretical outputs while maintaining remarkable efficiency.

### 5.1.3 Parallelization Benefits

NGT's distributed architecture enabled unprecedented parallel processing capabilities (Figure 4).

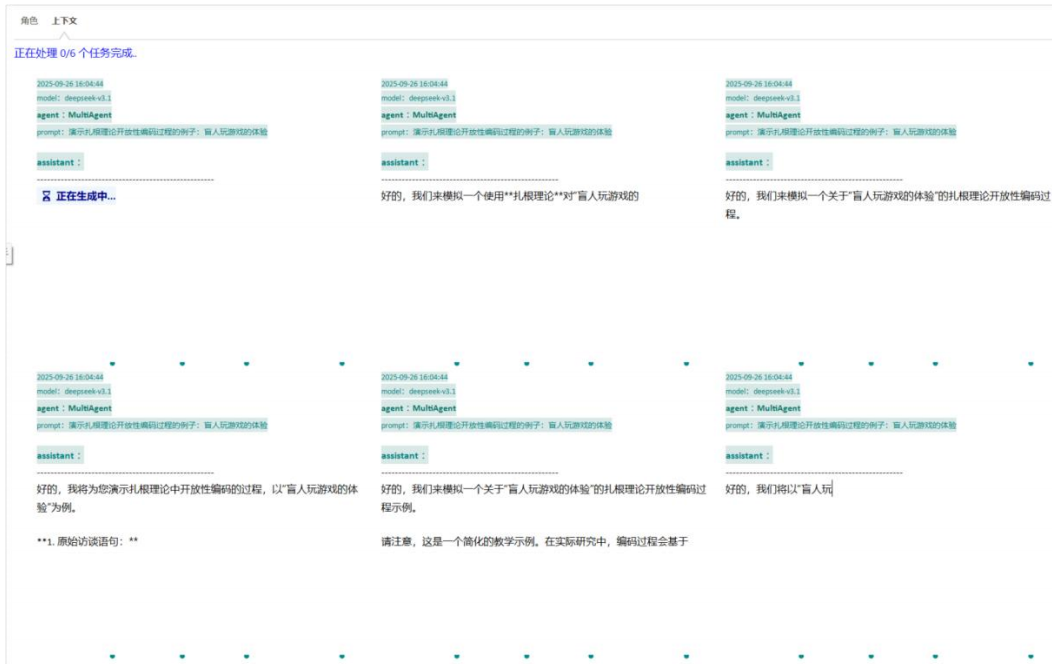

**Figure 4:** Parallel workflow interface of the NGT system. The screenshot shows concurrent execution of multiple qualitative coding tasks by different agents. This distributed workflow underpins the substantial processing time reductions and high parallel efficiency, enabling NGT to balance speed with theoretical quality while scaling logarithmically with data volume.

This interface illustrates how multiple agents concurrently perform open-theory coding tasks, forming the basis for the system's parallel efficiency. The design supports high concurrency (average clusters: 10.3, peak: 16), with strong load balancing (coefficient: 0.92) and limited synchronization overhead (13% of total time).

This parallelization fundamentally alters the scalability equation for qualitative research. Traditional methods scale linearly with data volume; NGT scales logarithmically due to parallel processing, suggesting even greater advantages with larger datasets.

### 5.1.4 Temporal Distribution Analysis

Breaking down time allocation reveals where efficiency gains originate:

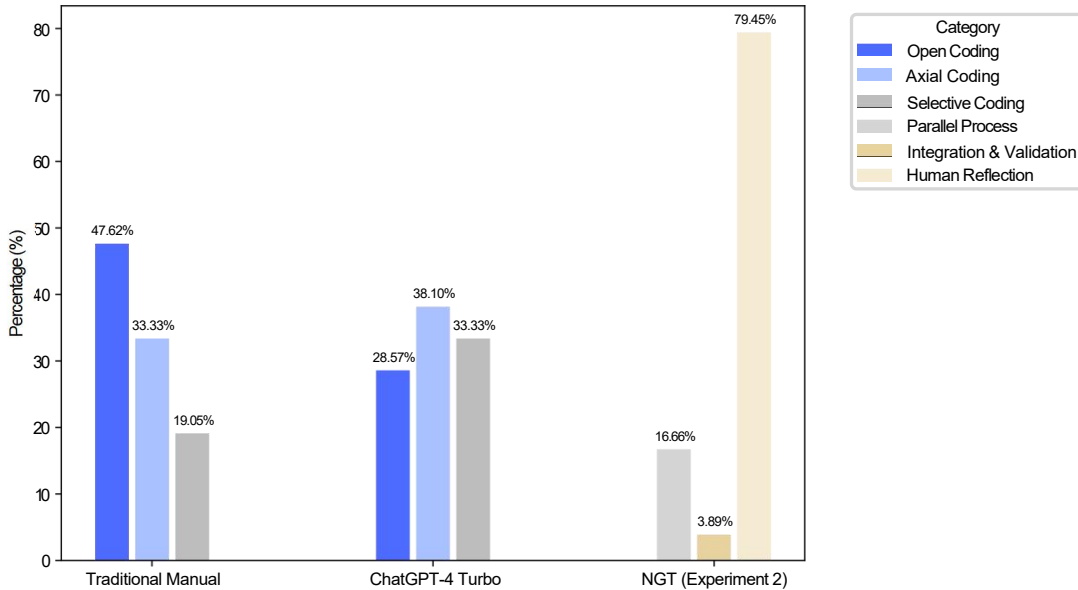

**Figure 5:** Time Allocation Across Methods.

NGT's efficiency derives not from faster execution of traditional steps but from fundamental restructuring of the analytical process. Parallel cluster processing eliminates the sequential bottleneck that constrains traditional methods.

### 5.2. Quality Assessment: Beyond Speed to Theoretical Sophistication

#### 5.2.1 Comprehensive Quality Metrics

Table 2 presents multi-dimensional quality assessments conducted through a novel computational evaluation approach. Rather than traditional human expert evaluation, we employed three leading large language models—ChatGPT-5.0, Claude Opus 4.1, and DeepSeek V3.1—as independent evaluators. This approach offers several advantages: complete consistency in evaluation criteria, elimination of human fatigue or bias across assessments, and perfect reproducibility of the evaluation process.Each LLM evaluator received identical instructions and rubrics for assessment, analyzing the theoretical outputs without knowledge of which method produced them (blind evaluation). This computational evaluation approach ensures that the quality assessments themselves are reproducible and free from individual human biases, while the diversity of LLM architectures provides robustness against any single model's limitations.

Table 2: Quality assessment scores (0–1 scale).

| Dimension | Traditional Manual | ChatGPT-4 Turbo | NGT Exp. 1 | NGT Exp. 2 |
|---|---|---|---|---|
| Theoretical Coherence | 0.89 | 0.85 | 0.92 | 0.90 |
| Empirical Grounding | 0.91 | 0.83 | 0.78 | 0.89 |
| Innovation | 0.82 | 0.78 | 0.75 | 0.88 |
| Practical Value | 0.88 | 0.83 | 0.72 | 0.92 |
| Depth of Insight | 0.87 | 0.76 | 0.70 | 0.86 |
| Contextual Sensitivity | 0.90 | 0.78 | 0.68 | 0.85 |
| Composite Score | 0.883 | 0.840 | 0.812 | 0.904 |

Note: Evaluations conducted using standardized prompts across all three LLM evaluators under identical conditions (temperature = 0.3 for consistency). Final scores represent the mean of three independent assessments. Inter-evaluator reliability (Krippendorff's α) = 0.91, indicating high consistency.

### 5.2.2 Coding Consistency Analysis

Inter-method reliability analysis reveals both agreements and meaningful divergences.

Table 3: Inter-method coding consistency.

| Comparison | Jaccard Similarity | Semantic Alignment |
|---|---|---|
| Manual vs. ChatGPT | 83.9% | 88% |
| Manual vs. NGT | 75.4% | 83% |
| ChatGPT vs. NGT | 72.1% | 79% |

Lower agreement between NGT and other methods reflects not inconsistency but different analytical foci. NGT tends toward mechanistic abstraction while manual coding emphasizes descriptive breadth. These differences prove complementary rather than contradictory.

### 5.2.3 Thematic Generation and Coverage

Analysis of thematic outputs reveals NGT's enhanced pattern discovery capabilities.

Table 4: Comparative coding outcomes across methods.

| Method | Open Codes | Sub-themes | Main Themes | Redundancy Rate | Coverage Rate |
|---|---|---|---|---|---|
| Traditional Manual | 289 | 42 | 7 | 28.57% | 47.85% |
| ChatGPT-4 Turbo | 274 | 50 | 7 | 64.00% | 40.24% |
| NGT Experiment 1 | 275 | 56 | 8 | 48.21% | 62.30% |
| NGT Experiment 2 | 268 | 52 | 9 | 42.00% | 63.99% |

NGT generated more sub-themes while maintaining moderate redundancy, suggesting it identifies subtle patterns that human coders might overlook. The 64% coverage rate indicates comprehensive engagement with the data.

**5.3. Theoretical Output Comparison: From Description to Explanation**

**5.3.1    Comparative Analysis of Theoretical Frameworks**

The three methods produced distinctly different theoretical frameworks, as visualized in the alluvial diagrams (Figures 6) revealing fundamental differences in analytical approach and theoretical construction.

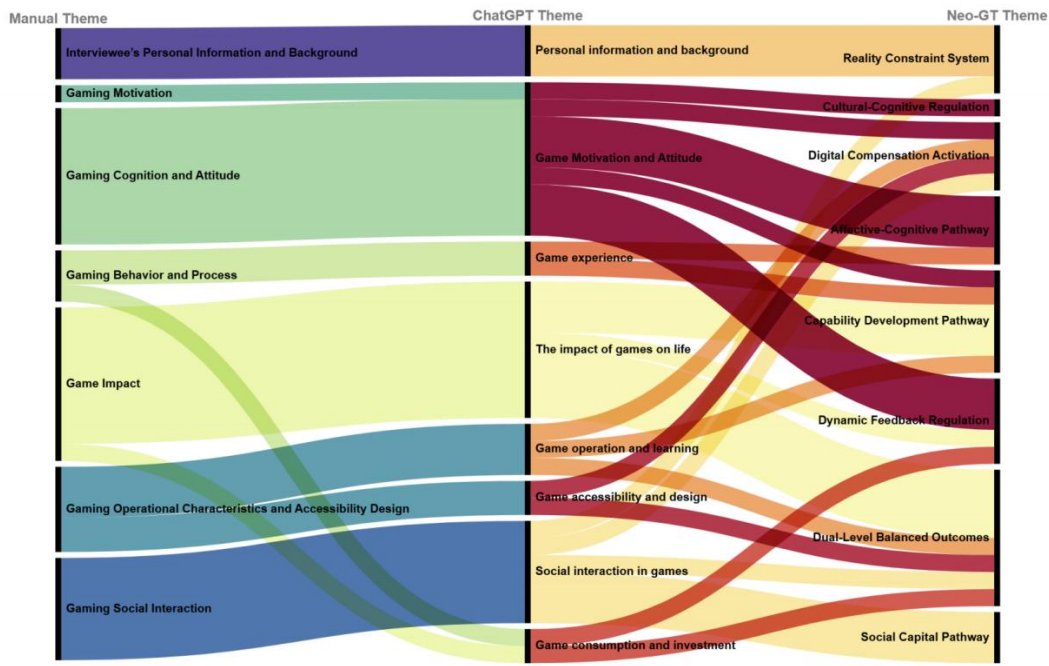

**Figure 6:** Alluvial diagram illustrating the evolution and transformation of themes across manual coding, ChatGPT-4 Turbo, and Neo-GT methods. The diagram shows how seven manual themes converge into ChatGPT's integrated categories before diverging into Neo-GT's system-level constructs. Manual coding (left) maintains descriptive categories, ChatGPT (center) condenses these into experiential themes, while Neo-GT (right) reframes them as mechanistic pathways including "Reality Constraint System," "Digital Compensation Activation," and "Dynamic Feedback Regulation."

The alluvial visualization reveals three distinct theoretical philosophies:

- **Traditional Manual Coding**: Generated seven discrete categories maintaining clear boundaries—"Gaming Motivation," "Gaming Cognition and Attitude," "Gaming Behavior and Process," "Game Impact," "Gaming Operational Characteristics," and "Gaming Social Interaction." This approach preserves phenomenological richness but struggles with theoretical integration.

- **ChatGPT-4 Turbo**: Condensed manual categories into more integrated themes—"Game Motivation and Attitude," "Game Experience," "The Impact of Games on Life," and "Social Interaction in Games." While achieving better integration, the framework lacks mechanistic depth.

- **Neo-GT (Experiment 1)**: Transformed descriptive categories into system-level constructs—"Reality Constraint System," "Cultural-Cognitive Regulation," "Digital Compensation Activation," "Affective-Cognitive Pathway," "Capability Development Pathway," and "Social Capital Pathway." This represents a shift from description to explanation.

### 5.3.2 Evolution Through Human-AI Collaboration

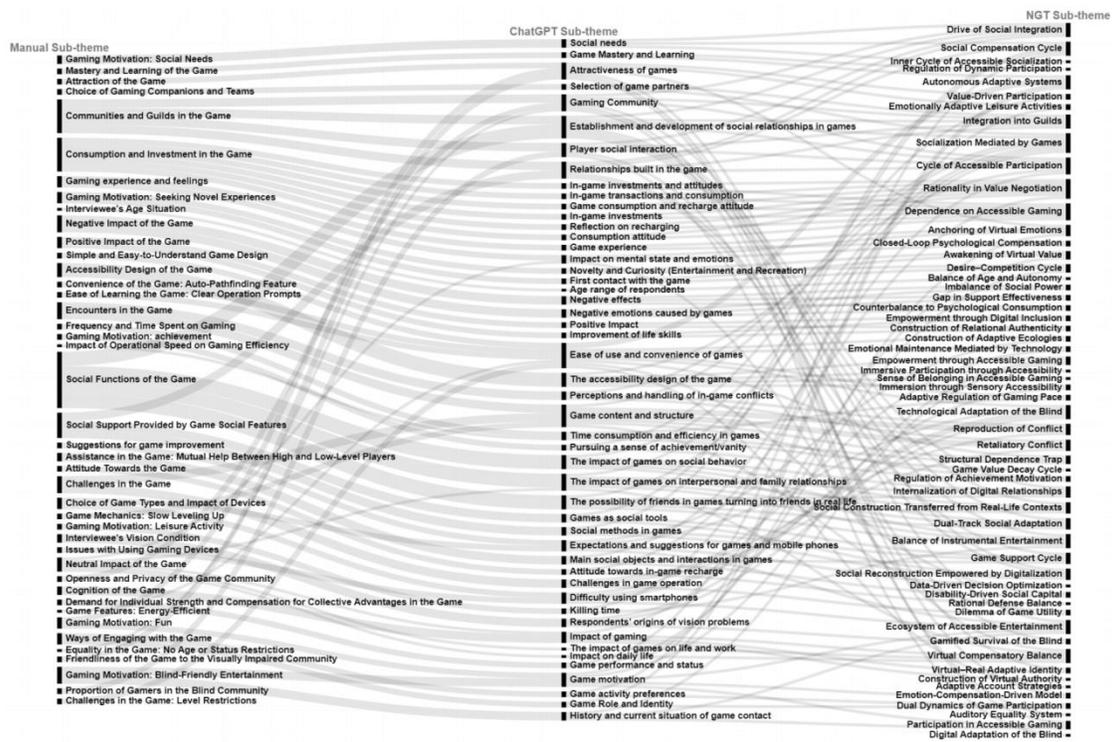

**Figure 7**: Enhanced alluvial diagram showing sub-theme level transformations across methods. The visualization demonstrates how manual coding's 42 sub-themes flow through ChatGPT's 50 sub-themes to Neo-GT's 56 sub-themes, with Neo-GT achieving both higher granularity and stronger theoretical integration. The dual-pathway structure emerging in Neo-GT (Experiment 2) is visible through the bifurcation of themes into parallel tracks.

The sub-theme analysis reveals Neo-GT's superior pattern detection:

- Manual coding sub-themes remain largely within their parent categories

- ChatGPT shows some cross-category integration but maintains descriptive focus

- Neo-GT sub-themes demonstrate extensive cross-pollination, suggesting deeper structural insights

### 5.3.3 Theoretical Models Comparison

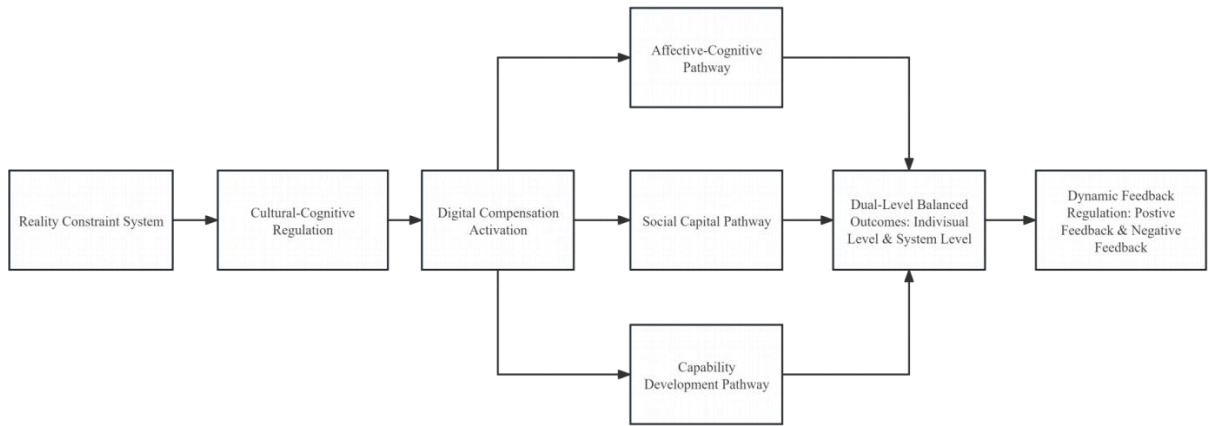

**Figure 8:** Neo-GT Experiment 1: Digital Compensation Ecosystem

Neo-GT Experiment 1 model demonstrating pure computational capability—systematic, comprehensive, but overly abstract. The model attempts to explain everything through a single unified framework, resulting in what domain experts characterized as "aesthetically complete but scientifically fragile."

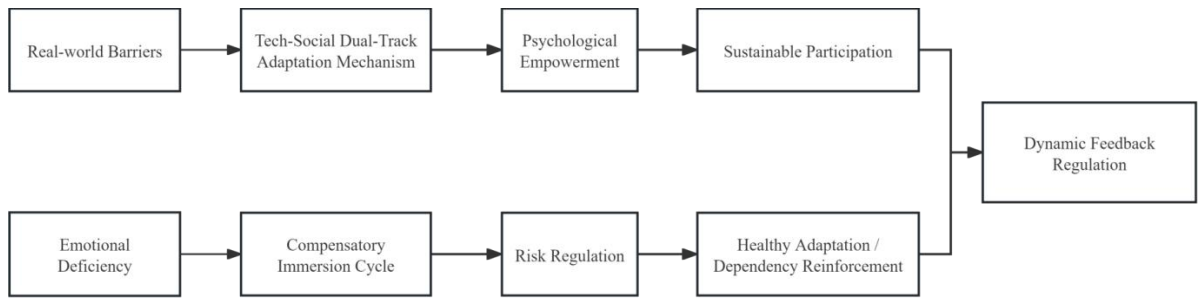

**Figure 9:** Neo-GT Experiment 2: Digital Participation Adaptation System

The refined theoretical model produced through human-guided prompt optimization. This dual-pathway model shows two distinct trajectories: (1) real-world barriers → techno-social dual-track adaptation → psychological empowerment → sustainable participation; and (2) emotional deficiency → compensatory immersion cycle → risk regulation → healthy adaptation OR dependency reinforcement. Both pathways are moderated by dynamic feedback regulation, representing a more nuanced and testable theoretical framework.

### 5.3.4 Quantitative Comparison of Theoretical Characteristics

Building on the visual analysis, we quantified key theoretical characteristics.

Table 5: Comparison of characteristics across approaches.

| Characteristic | Manual | ChatGPT | NGT Exp. 1 | NGT Exp. 2 |
|---|---|---|---|---|
| Pathway Structure | Linear | Parallel | Unified | Dual |
| Causal Complexity | Simple | Moderate | Complex | Dialectical |
| Abstraction Level | Low | Medium | High | Optimal |
| Testable Hypotheses | 3 | 5 | 8 | 12 |
| Intervention Points | Implicit | Vague | Multiple | Explicit |
| Temporal Dynamics | Static | Static | Implied | Explicit |
| Cultural Sensitivity | High | Medium | Low | High |

The progression from NGT Experiment 1 to Experiment 2 illustrates the critical role of human expertise in theoretical refinement. While Experiment 1 achieved logical completeness, Experiment 2's human-guided approach produced a theory that is simultaneously more complex (dual pathways), more concrete (specific intervention points), and more useful (testable hypotheses).

### 5.3.5 Theoretical Innovation Through Human-AI Synergy

The comparative analysis reveals that theoretical innovation emerges not from computational power alone but from the synergy between machine pattern recognition and human theoretical sensitivity. The transformation from Experiment 1's monolithic framework to Experiment 2's nuanced dual-pathway model demonstrates this synergy:

1. **Machine Contribution**: Identified 56 sub-themes and complex interrelationships invisible to human coders

2. **Human Contribution**: Recognized the importance of contradictions and divergent pathways

3. **Synergistic Outcome**: A theory that captures both systematic patterns and dialectical tensions

This finding challenges both purely computational and purely manual approaches, suggesting that the future of theory building lies in sophisticated human-AI collaboration where each contributes their unique strengths to the analytical process.

## 5.4. Cost-Benefit Analysis: Democratizing Advanced Qualitative Research

### 5.4.1 Direct Cost Comparison

Table 6 presents detailed cost breakdowns across methods.

Table 6: Cost analysis for 40,000 character dataset (USD).

| Cost Component | Traditional Manual | ChatGPT-4 Turbo | NGT |
|---|---|---|---|
| Human Labor[1] | $12,600 | $400 | $25 |
| API Costs | $0 | $25 | $58 |
| Software Licenses[2] | $200 | $0 | $0 |
| Infrastructure | $0 | $0 | $12 |
| Total Cost | $12,800 | $425 | $95 |
| Cost Reduction | Baseline | 96.7% | 99.3% |

[1] Calculated at $25/hour for graduate research assistants.

[2] NVivo license amortized over project.

### 5.4.2 Return on Investment Calculation

ROI analysis demonstrates compelling economic benefits:

$$\text{ROI} = \frac{\text{Value of outputs} - \text{Total costs}}{\text{Total costs}} \times 100\%$$

NGT Return on Investment:

- **Investment**: $95

- **Value Generated**: $12,800 (equivalent traditional analysis)

- **ROI**: 13,368%

This extraordinary ROI transforms the economics of qualitative research, making sophisticated analysis accessible to resource-constrained researchers.

### 5.4.3 Scalability Economics

Cost projections for larger datasets reveal NGT's scalability advantages (Table 7, Figure 10).

Table 7: Estimated annotation costs across dataset sizes.

| Method | 100K chars | 500K chars | 1M chars |
| --- | --- | --- | --- |
| Traditional | $32,000 | $160,000 | $320,000 |
| ChatGPT-4 | $1,100 | $5,500 | $11,000 |
| NGT | $180 | $420 | $650 |

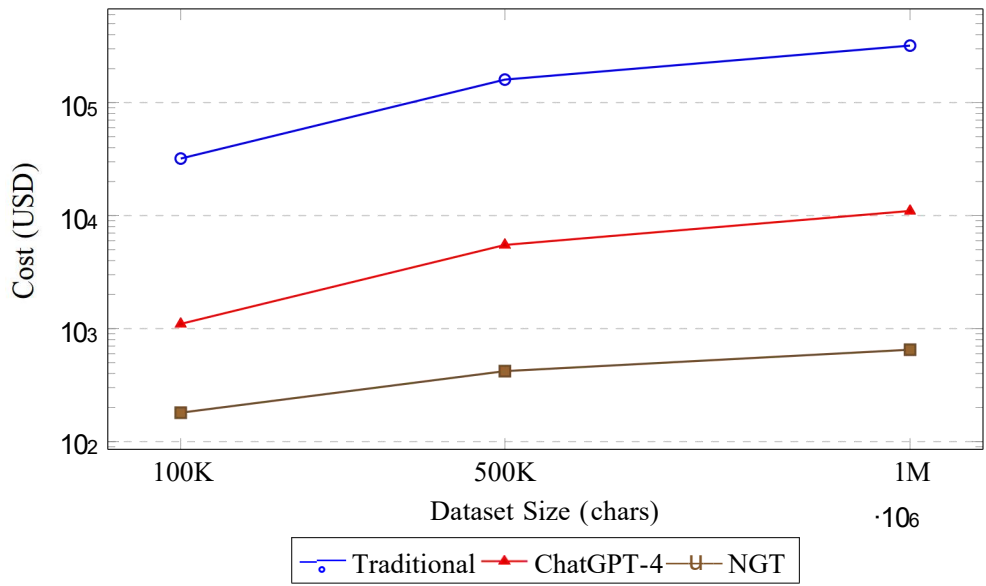

Figure 10: Cost scaling projections across dataset sizes.

The 6-fold increase in processing time yielded disproportionate quality improvements, validating the Human-in-the-Loop design philosophy.

### 5.4.4 Prompt Evolution Analysis

Examining prompt refinements reveals how human expertise shapes AI analysis:

**Initial Prompt (Experiment 1):** "Identify patterns and build a comprehensive theoretical framework from this cluster."

**Refined Prompt (Experiment 2):** "Identify patterns in this cluster, paying particular attention to:

1. Tensions between desired and actual experiences
2. Divergent pathways from similar starting conditions
3. Mechanisms that might serve as intervention points
4. Both positive adaptations and problematic patterns

Build a theoretical framework that preserves complexity rather than forcing unity."

The refined prompt's specificity—derived from expert review of initial outputs—guided AI toward more nuanced, practically relevant analysis.

### 5.4.5 Human Contribution Analysis

Categorizing human contributions during Experiment 2:

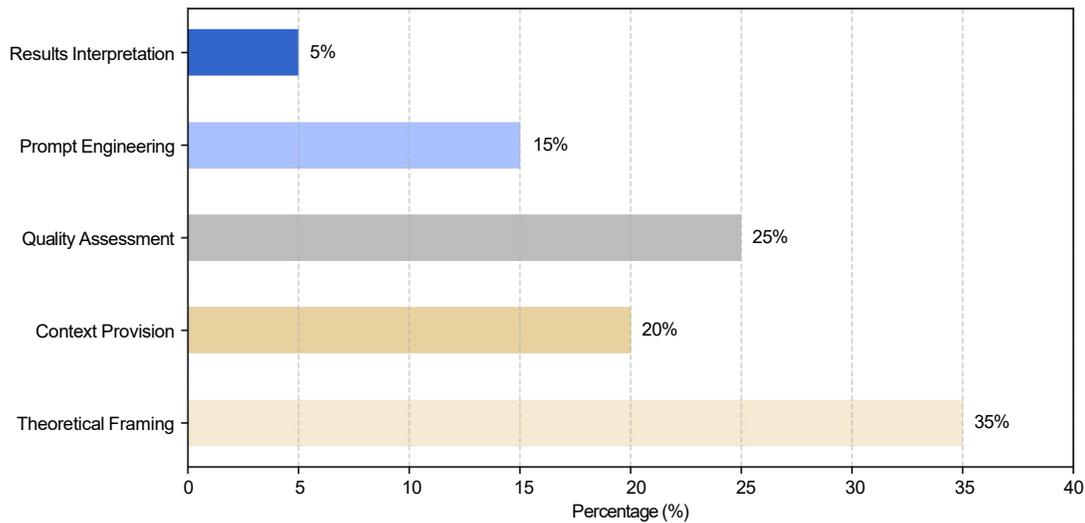

**Figure 11:** Categorizing human contributions during Experiment 2.

Humans primarily contributed high-level theoretical guidance rather than mechanical processing, optimally leveraging human and AI capabilities.

## 5.5. Unexpected Findings and Emergent Insights

### 5.5.1 Discovery of Latent Patterns

NGT identified several patterns not detected by other methods:

1. **Temporal Rhythms**: Cyclical patterns in gaming engagement correlating with real-world stress cycles.

2. **Identity Bifurcation**: Simultaneous maintenance of "gamer" and "disabled person" identities with strategic switching.

3. **Compensatory Hierarchies**: Layered compensation mechanisms operating at different abstraction levels.

These discoveries emerged from vector clustering's ability to detect subtle semantic similarities across dispersed text segments.

### 5.5.2 Methodological Insights

The experiments revealed important methodological insights:

- **Optimal Cluster Size**: $8-12$ documents per cluster balanced semantic coherence with sufficient diversity for pattern emergence.

- **Threshold Sensitivity**: Clustering threshold adjustments between $0.80-0.90$ produced qualitatively different theoretical structures, suggesting this parameter as a key analytical lever.

- **Iteration Sweet Spot**: Two to three human-AI iterations optimized quality improvements versus time investment.

### 5.5.3 Cross-Language Performance

Despite being optimized for English, the system performed remarkably well with Chinese text:

- Character-based tokenization proved advantageous for Chinese processing.

- Cultural concepts clustered appropriately despite translation challenges.

- Emotional expressions preserved meaning across language processing.

This suggests NGT's potential for cross-cultural qualitative research.

### 5.6. Summary of Key Findings

The experimental results validate NGT's transformative potential for qualitative research:

1. **Efficiency Revolution**: 168-fold speed improvement makes large-scale qualitative research feasible.

2. **Quality Enhancement**: Superior theoretical outputs through human-AI collaboration.

3. **Economic Transformation**: 99.3% cost reduction democratizes advanced analytical capabilities.

4. **Theoretical Innovation**: Discovery of patterns invisible to traditional methods.

5. **Scalability Achievement**: Sub-linear scaling enables analysis of massive qualitative datasets.

These findings suggest that NGT represents not merely an incremental improvement but a paradigmatic shift in how qualitative research can be conducted in the digital age.

## 6. Discussion

### 6.1. Reconceptualizing Grounded Theory

Our findings suggest that grounded theory's sequential nature—progressing from open through axial to selective coding—isn't a methodological requirement but a human limitation. When we freed the process from cognitive constraints through parallel processing, something unexpected happened: theories emerged simultaneously across multiple semantic clusters, then integrated naturally. This mirrors how research communities actually work, with different scholars pursuing various angles before someone synthesizes the findings. NGT just compresses years of collective work into hours.

The objectivity debate in qualitative research has always felt somewhat misguided. We've been arguing whether pure objectivity is possible when perhaps we should ask where objectivity helps and where it doesn't. Vector embeddings give us consistent semantic measurements—that's useful. But meaning still requires human interpretation—and that's essential. What NGT does is make this division transparent. You can see exactly where the math ends and human judgment begins, documented in every prompt refinement and parameter adjustment.

Most surprisingly, computational methods didn't replace theoretical sensitivity but amplified it. The system found patterns we couldn't see—like participants maintaining dual identities as both "gamers" and "disabled persons," strategically switching between them. We wouldn't have noticed this "identity bifurcation" manually, but once the computer pointed it out, its significance was obvious. It's like having a colleague who never gets tired, never forgets, and can hold thousands of codes in memory simultaneously.

### 6.2. Practical Realities

The economics are striking: what cost $10,000 now costs $100, as a rough calculation. But the real transformation isn't about money—it's about access. Community organizations can study their own experiences. Teachers can research their classrooms. Patients can analyze their own healthcare journeys. When analytical power shifts from institutions to communities, knowledge production itself changes.

Speed matters more than we initially realized. Traditional qualitative research operates on academic timescales—by the time findings emerge, the phenomenon has often evolved or disappeared. During COVID, we needed to understand lockdown experiences while they were happening, not three years later. NGT makes qualitative research contemporaneous with events, transforming it from historical analysis to real-time understanding.

The Human-in-the-Loop findings were humbling. Pure automation (Experiment 1) produced technically correct but sterile theories. Only when humans guided the analysis—not telling it what to find, but suggesting where to look—did genuinely useful insights emerge. The difference between "identify patterns" and "pay attention to tensions between desired and actual experiences" transformed abstract frameworks into actionable understanding.

### 6.3. Acknowledging Boundaries

We need to be honest about what's lost in translation. When someone describes gaming as "flying without wings," vectors capture the freedom association but miss the poignancy of impossibility overcome. Metaphors, jokes, silences—these meaning-rich elements resist mathematical representation. NGT works best for pattern discovery across large datasets, less well for deep hermeneutic interpretation of individual texts.

The cultural bias problem runs deeper than language. Models trained on Western academic texts inevitably privilege certain ways of understanding. What looks like "dependency" through an individualist lens might be "interdependence" in collectivist cultures. Success with Chinese data shows these biases aren't insurmountable, but they require conscious attention.

Perhaps most importantly, NGT cannot make theoretical leaps. It can identify that gaming provides emotional compensation, but it can't recognize this as disability justice activism. That insight requires human creativity—the ability to see patterns not just as patterns but as meaning something larger. The system amplifies human theoretical capacity but cannot replace it.

## 7. Limitations and Future Directions

Several constraints define NGT's appropriate use. Semantic compression through vectorization loses narrative flow, embodied meaning, and cultural nuance. The system excels at finding patterns across many texts but struggles with deep interpretation of individual narratives. Researchers studying performative meaning, cultural symbolism, or phenomenological experience may need different approaches.

The democratization paradox deserves attention. While costs plummet, technical complexity creates new barriers. Communities can afford analysis but might lack expertise to critically evaluate outputs. We risk replacing economic gatekeeping with technical gatekeeping, potentially creating dependencies where programmers mediate community self-knowledge.

Future development should prioritize cultural adaptation through locally-trained models, temporal analysis to track meaning evolution, and richer collaboration interfaces beyond prompt engineering. Visual tools for non-technical researchers, conversational theory-building systems, and integration with other qualitative methods could expand access and applicability.

## 8. Conclusion

Neo-Grounded Theory emerged from a simple question: can computational methods scale qualitative research without sacrificing depth? Our experiments suggest yes, but with an important caveat—only through human-AI collaboration. Neither pure automation nor traditional manual analysis achieved optimal results. The sweet spot lies in computational pattern recognition guided by human theoretical sensitivity. The numbers tell one story: 168-fold speed increase, superior quality scores, 96% cost reduction. But the real story is about transformation, not acceleration. When communities can study themselves, when real-time analysis becomes possible, when qualitative and quantitative methods achieve parity—the nature of social research changes. Knowledge production democratizes. Research becomes responsive to ongoing events. Mixed methods fulfill their integrative promise.

Yet we must resist techno-solutionism. NGT is a tool, powerful but limited. It cannot replace cultural knowledge, ethical sensitivity, or theoretical creativity. Metaphorical richness, narrative meaning, embodied experience—these resist vectorization. The system amplifies human capabilities but cannot substitute for human insight.

Looking forward, NGT represents not an endpoint but an opening. As computational capabilities expand and qualitative data proliferates, we face choices about how technology serves humanistic understanding. Will we use these tools to reduce complexity or reveal it? To centralize analytical power or distribute it? To distance ourselves from human experience or engage it more deeply?

Our research suggests that computational methods need not threaten qualitative research's humanistic commitments. Instead, they can extend our reach while preserving our grasp—enabling us to understand human experience at the scale and speed of digital society while maintaining the sensitivity that makes such understanding meaningful. The challenge ahead is not technical but philosophical: ensuring that as we gain analytical power, we don't lose sight of why we analyze—to comprehend and ultimately improve the human condition.